\documentclass[10pt,twocolumn,letterpaper]{article}
\usepackage[pagenumbers]{cvpr}
\usepackage{amsmath,amssymb,graphicx,booktabs,array,xcolor,tabularx,enumitem}
\definecolor{cvprblue}{rgb}{0.21,0.49,0.74}
\usepackage[pagebackref,breaklinks,colorlinks,allcolors=cvprblue]{hyperref}
\def\confName{CVPR}
\def\confYear{2027}

\title{VisualTalk-Bench: Evaluating Question-Driven Avatar-Led Video Explanations}

\author{
Xin Zhang$^{1}$, Tujie Xu$^{2}$, Zhiyi Chen$^{2}$,\\
Xiaodong Zhang$^{1}$, Sicheng Yang$^{3}$, Minglei Li$^{2}$\\[0.5em]
$^{1}$University of Science and Technology of China\\
$^{2}$Leadde AI\qquad $^{3}$Tsinghua University
}
\hypersetup{
  pdfauthor={Xin Zhang, Tujie Xu, Zhiyi Chen, Xiaodong Zhang, Sicheng Yang, Minglei Li},
  pdftitle={VisualTalk-Bench: Evaluating Question-Driven Avatar-Led Video Explanations}
}
\begin{document}
\maketitle
\raggedbottom

\begin{abstract}
Advances in video generation and avatar synthesis enable new ways to answer user questions through visual explanations. This task requires a system to formulate a correct answer, translate relevant information into visual material, and deliver the explanation through an avatar's speech and actions. Existing benchmarks rarely assess whether avatars point to the visual content being discussed. We introduce VisualTalk-Bench for question-driven avatar-led video explanations in finance and Chinese and English education. It includes 800 curated reference videos and 100 independent evaluation requests, assessing Content Factuality, Speech--Visual Consistency, Visual Quality, Motion Smoothness, and Avatar--Content Pointing. Yet end-to-end video generators often struggle to align gestures with specific visual targets. To explore this challenge, we develop Avatar-O Agent, a modular reference implementation with explicit interaction planning. Experiments across six agents show that factual precision and visual quality do not guarantee accurate pointing. A 26-participant study comparing four agents finds that Avatar-O Agent receives the highest average ratings for visual guidance, perceived pointing accuracy, and explanation effectiveness. We will release the data and code.
\end{abstract}

\section{Introduction}
Recent advances in video generation and avatar synthesis are enabling systems to answer user questions through avatar-led visual explanations. Existing work generates narrated presentations from source documents~\cite{mpresenter,presentagent,paper2video}, while query-driven approaches use the user's request to guide presentation creation~\cite{presentagent2}. This task brings together question answering and video creation: the system must determine what information is needed, choose a suitable visual representation, and deliver the explanation to the viewer.

Evaluating these systems requires understanding how well the generated videos explain the requested information. Existing evaluations address several aspects of this problem, including visual quality~\cite{vbench,evalcrafter}, conceptual correctness~\cite{eduvqa,eduvideobench}, and gesture naturalness and speech appropriateness~\cite{genea}. However, these measures provide limited insight into how the avatar interacts with the displayed content. An effective explanation also depends on whether the avatar's narration and actions are coordinated with the information shown on screen. This interaction remains insufficiently assessed by existing evaluation protocols.

An avatar may, for example, state a correct value and display an accurate chart while pointing to the wrong bar. The error lies in how the explanation connects speech, action, and visual evidence. Detecting it requires identifying the intended element and examining the avatar's action at the relevant moment. This becomes more challenging when the visual material is generated by the system itself. The same answer can be expressed through different charts, diagrams, or tables, with the relevant elements appearing in different locations. Evaluation must therefore locate the relevant content in each system's own video and assess the avatar's interaction with it in that context.

\begin{figure}[!t]
\centering
\includegraphics[trim=22bp 15bp 22bp 18bp,clip,width=\columnwidth]{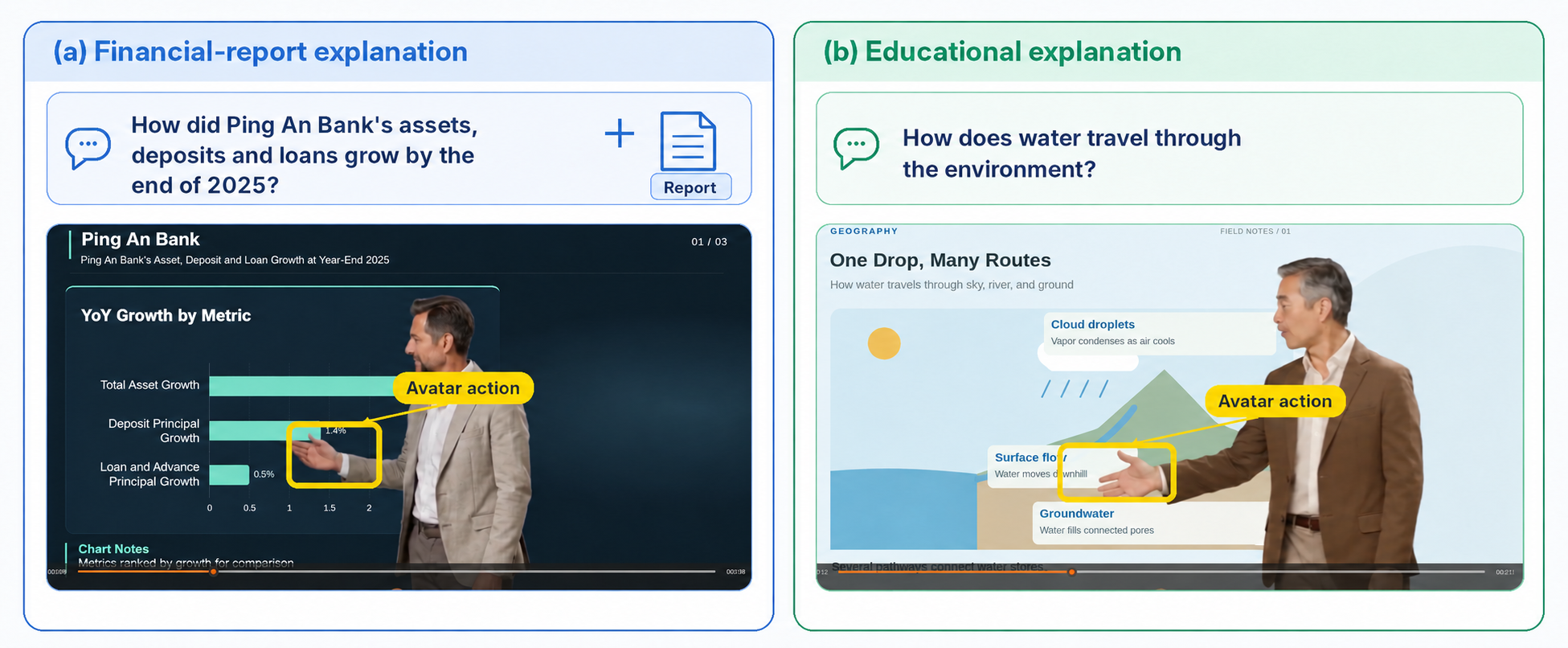}
\caption{Two VisualTalk-Bench tasks. (a) The avatar uses a chart to explain growth in Ping An Bank's assets, deposits, and loans from the supplied report. (b) Given a question about the water cycle, the avatar illustrates how water moves through the environment. Yellow boxes highlight the presenters' gestures.}
\label{fig:overview}
\end{figure}

We introduce \textbf{VisualTalk-Bench}, an architecture-agnostic benchmark for question-driven avatar-led visual explanations. Figure~\ref{fig:overview} illustrates two representative task settings: source-grounded financial-report explanations and document-free educational explanations, both delivered through avatar-led videos. It evaluates the delivered video, whether produced by a modular agent or an end-to-end generator. An evaluator first resolves a spoken reference in the system's own frames, then examines the gesture at that moment. Separating these decisions prevents the presenter's hand position or a planned gesture from determining the intended target.

Our reference implementation, \textbf{Avatar-O Agent}, links narration to rendered objects before video synthesis. We compare it with five other agent configurations under common requests and use viewer judgments to examine how the resulting explanations are perceived.

Our contributions are threefold:
\begin{itemize}[leftmargin=*,nosep]
\item \textbf{Task and benchmark.} We introduce VisualTalk-Bench for question-driven avatar-led visual explanation generation, with 800 manually screened reference video examples across source-grounded financial explanations and Chinese and English educational explanations. An independently constructed set of 100 evaluation requests supports the system comparison in this paper. We will make the full dataset publicly available and open-source all of our benchmark construction, agent orchestration, and evaluation code.
\item \textbf{Interaction-aware evaluation.} We evaluate five dimensions of video explanations, with particular emphasis on avatar--content pointing assessed from final videos independently of generation-time cues.
\item \textbf{Initial system study.} We provide Avatar-O Agent as a modular realization of the task and analyze six agent configurations under the same benchmark requests. The results reveal differences in content factuality, speech--visual consistency, and pointing accuracy.
\end{itemize}

\section{Related Work}

\noindent\textbf{Presentation and explanatory video generation.}
Slide-generation systems plan and edit layouts~\cite{pptagent}, while PresentAgent and mPresenter couple document-derived slides with narration~\cite{presentagent,mpresenter}. PaperTalker adds a talking head and narration-aligned cursor guidance to presentations of scientific papers~\cite{paper2video}. PresentAgent-2 uses queries to guide presentation creation and supports audience questions~\cite{presentagent2}. Programmatic approaches generate explanatory animations for mathematical and educational topics~\cite{theorem,code2video}. VisualTalk-Bench asks for a video answering a particular question; the source may be a financial report or, for educational questions, no document at all.

\noindent\textbf{Generated video and content evaluation.}
Video benchmarks assess perceptual quality, compositional relations, physical plausibility, and scientific reasoning~\cite{vbench,evalcrafter,t2vcomp,videophy,scivbench}. Educational benchmarks examine the validity of concepts and instructional content~\cite{eduvqa,eduvideobench}. Questions about a completed presentation can also test which information reaches the viewer~\cite{presentagent,mpresenter,paper2video}. Chart understanding and generation connect data to visual representations through question answering, summarization, and chart reconstruction~\cite{chartqa,plotqa,charttext,opencqa,lida,chartmimic}. Here the system creates its own visual evidence, so an evaluator must determine whether the relevant fact is present and whether narration refers to it at the appropriate time.

\noindent\textbf{Avatar gestures and visual grounding.}
Speech-driven portrait and gesture synthesis target lip, face, hand, and body motion~\cite{wav2lip,sadtalker,beat,emage}. GENEA measures gesture naturalness and speech appropriateness~\cite{genea}, while StreamEval assesses execution of requested gestures~\cite{streameval}. Audiovisual synchronization measures whether visible articulation follows the speech signal~\cite{syncnet}; a synchronized presenter can still gesture toward an unrelated object. Other work links narration or gestures to chart elements~\cite{dataplayer,visconductor} and identifies targets from language and pointing~\cite{yourefit,deepoint}. In a multimedia lesson, Li et al.~\cite{li2026embodied} found that color cues attracted the most attention to relevant material, while learners in the embodied-agent condition achieved the strongest learning outcomes. This motivates examining embodied guidance separately from visual highlighting.

\begin{table*}[!t]
\centering
\caption{Evaluation scope of representative benchmarks and protocols. A check (\(\checkmark\)) marks an explicit target, a triangle (\(\triangle\)) marks partial or indirect coverage, and a dash (\textemdash) marks a target outside the stated evaluation. Content validity includes factual or conceptual accuracy and measures of information transfer.}
\label{tab:related-benchmarks}
\small
\setlength{\tabcolsep}{4pt}
\renewcommand{\arraystretch}{1.12}
\begin{tabularx}{\linewidth}{@{}l>{\raggedright\arraybackslash}Xccccc@{}}
\toprule
\textbf{Benchmark or protocol} & \textbf{Evaluation setting} &
\shortstack{\textbf{Visual}\\\textbf{quality}} &
\shortstack{\textbf{Content}\\\textbf{validity}} &
\shortstack{\textbf{Speech--}\\\textbf{visual}} &
\shortstack{\textbf{Avatar}\\\textbf{gesture}} &
\shortstack{\textbf{Avatar--Content}\\\textbf{Pointing}}\\
\midrule
VBench~\cite{vbench} & Text-to-video generation & \(\checkmark\) & \textemdash & \textemdash & \textemdash & \textemdash \\
T2V-CompBench~\cite{t2vcomp} & Compositional video generation & \(\triangle\) & \textemdash & \textemdash & \textemdash & \textemdash \\
Sci-VBench~\cite{scivbench} & Scientific video generation & \(\checkmark\) & \(\checkmark\) & \textemdash & \textemdash & \textemdash \\
\addlinespace[2pt]
EduAVQABench~\cite{eduvqa} & Concept-driven educational videos & \(\checkmark\) & \(\checkmark\) & \textemdash & \textemdash & \textemdash \\
EduVideoBench~\cite{eduvideobench} & Classroom video generation & \(\checkmark\) & \(\checkmark\) & \(\triangle\) & \textemdash & \textemdash \\
PresentEval~\cite{presentagent} & Document-based presentations & \(\checkmark\) & \(\checkmark\) & \(\triangle\) & \textemdash & \textemdash \\
Paper2Video~\cite{paper2video} & Scientific paper presentations & \(\triangle\) & \(\checkmark\) & \(\triangle\) & \textemdash & \textemdash \\
PresentEval (PresentAgent-2)~\cite{presentagent2} & Query-led presentations and audience QA & \(\triangle\) & \(\checkmark\) & \(\triangle\) & \textemdash & \textemdash \\
\addlinespace[2pt]
GENEA Challenge~\cite{genea} & Speech-driven gestures & \textemdash & \textemdash & \textemdash & \(\checkmark\) & \textemdash \\
StreamEval~\cite{streameval} & Timed interactive video tasks & \(\checkmark\) & \(\triangle\) & \textemdash & \(\checkmark\) & \textemdash \\
\midrule
\textbf{VisualTalk-Bench (ours)} & Avatar-led video answers & \(\checkmark\) & \(\checkmark\) & \(\checkmark\) & \(\checkmark\) & \(\checkmark\) \\
\bottomrule
\end{tabularx}
\par\smallskip
\raggedright\footnotesize
\emph{Speech--visual} measures whether the narration agrees with the displayed material.
\emph{Avatar gesture} covers gesture quality or execution.
\emph{Avatar--Content Pointing} assesses whether the avatar makes a directed gesture during a spoken reference and whether that gesture identifies the intended visual element.
\end{table*}

Table~\ref{tab:related-benchmarks} summarizes the evaluation scope of representative benchmarks across five dimensions. General video-generation benchmarks primarily assess visual quality, while scientific, educational, and presentation-oriented evaluations additionally consider content validity and, in some cases, speech--visual consistency. Gesture-focused evaluations examine motion quality or gesture execution, but do not explicitly assess whether an observed gesture identifies the visual element referenced in narration. Consequently, existing protocols address individual aspects of explanatory video generation without jointly evaluating factual validity, cross-modal correspondence, and object-level avatar--content interaction. VisualTalk-Bench addresses this gap by evaluating these dimensions together, with particular emphasis on grounding observed pointing gestures in independently identified visual referents.

\section{VisualTalk-Bench}
\label{sec:task}
\subsection{Task and System Inputs}
\label{sec:taskinput}
VisualTalk-Bench evaluates a system's ability to answer a single user request through an avatar-led explanatory video. Let \(q\) denote the question, \(D\) the optional source document, \(\ell\) the target language, and \(a\) the full-body avatar reference image. For financial questions, \(D\) contains selected pages of a report; for educational questions, \(D=\varnothing\). A system \(S\), whether modular or end-to-end, returns an ordered sequence of scene videos,
\[
V_q=S(q,D,\ell,a)=(v_{q,1},\ldots,v_{q,K_q}).
\]

The scenes jointly form one explanation. The presenter should communicate the information relevant to the question, display appropriate visual material, and coordinate gestures with the objects or relations mentioned in speech. Here, \emph{interaction} refers specifically to the presenter's interaction with the visual content \textbf{within a generated answer}; the task does not require live dialogue, interruptions, or multi-turn control.

Evaluation observes the delivered video independently of internal plans or controls. The curated corpus provides reference examples, while a separate evaluation set supports comparisons across systems. Reference answers and evaluator-side annotations are withheld from generators. The reference example illustrates one way to convey the required relationship. Other layouts can be equally valid, so the evaluator must locate the evidence in each generated answer.

\subsection{Reference Corpus Construction and Quality Control}
\label{sec:data}

\begin{figure}[!t]
\centering
\includegraphics[width=\columnwidth]{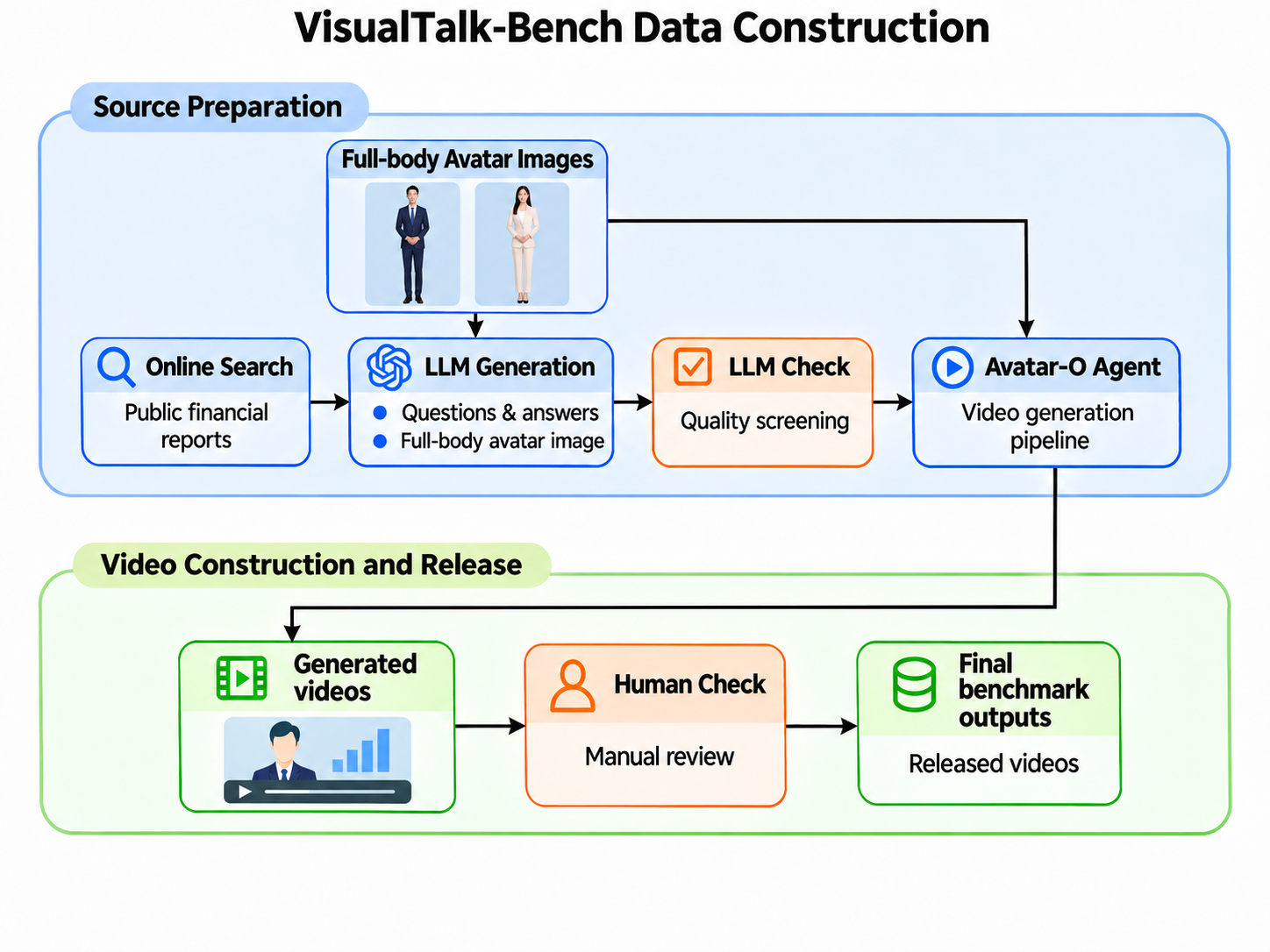}
\caption{Reference-video corpus construction. Full-body avatar reference images are prepared separately. Generated videos undergo automatic checks and human review before inclusion in the corpus.}
\label{fig:data-construction}
\end{figure}

Figure~\ref{fig:data-construction} illustrates the construction pipeline of our curated reference-video corpus. We first collect publicly available financial reports and curate educational topics in Chinese and English. An LLM generates questions, reference answers, and explanation plans, while full-body avatar reference images are prepared separately. An LLM-based quality check screens the generated content for factual grounding and suitability before video generation. We use Avatar-O Agent, a modular reference implementation of the task-to-video pipeline described in Section~\ref{sec:systems}, to generate candidate videos from the screened content. The agent prepares the narration scripts and scene images, synthesizes narration audio, and generates presenter videos with the planned interaction cues. It combines the presenter videos with the scene images and narration audio to produce candidate explanations, which subsequently undergo automatic checks and human review.

For financial questions, we retain supporting passages and page numbers from 100 publicly disclosed financial reports collected from CNINFO~\cite{cninfo}, covering 56 companies (50 annual reports for 2025 and 50 half-year reports for 2026). Educational topics follow subject catalogs~\cite{scienceqa,ai2d}.

\noindent\textbf{Quality Screening.} The construction process involved over 1,500 candidate examples. Automatic checks examined narration--visual correspondence, audio duration, and planned interaction intervals. Two independent human reviewers evaluated the candidates for content correctness and explanation suitability, including factual grounding in financial reports and the validity of educational explanations. A candidate was retained only when both reviewers approved it. This screening process yielded 800 complete reference-video examples, comprising 300 financial, 300 Chinese educational, and 200 English educational examples.

\subsection{Dataset Composition and Independent Evaluation Set}
\label{sec:subset}
Each reference video includes its request, scene-level narration, visual assets, audio, and any source evidence. We independently constructed 100 evaluation requests disjoint from the 800 corpus examples. The 20 financial requests (ten annual, ten half-year) cover 16 reports from 11 companies. The 40 Chinese and 40 English educational requests include ten and five questions per subject, respectively. Subject differences limit comparisons between languages.

Financial answers have three ordered scenes and educational answers one, totaling 140 scenes per agent and 840 across six configurations. We score every output, including flawed charts and gestures. Reference videos are excluded from the reported results.

\subsection{Evaluation Protocol}
\label{sec:eval}
VisualTalk-Bench evaluates five complementary dimensions of generated explanations, summarized in Table~\ref{tab:eval_dimensions}. These dimensions are reported separately without aggregation into a single overall score. We describe the scoring criteria below.

\begin{table}[!ht]
\centering
\caption{Overview of VisualTalk-Bench evaluation dimensions. Each dimension assesses a distinct aspect of the generated explanation; no overall score is computed.}
\label{tab:eval_dimensions}
\small
\setlength{\tabcolsep}{3pt}
\begin{tabularx}{\columnwidth}{@{}>{\raggedright\arraybackslash}p{.25\columnwidth}>{\raggedright\arraybackslash}X>{\raggedright\arraybackslash}p{.28\columnwidth}@{}}
\toprule
Evaluation Dimension & Evaluation Focus & Reported Metric \\
\midrule
Content Factuality & Factual validity of spoken and visual claims & Speech / Visual Factual Precision~$\uparrow$ \\
Speech--Visual Consistency & Agreement between narration and displayed content & Consistency Score~$\uparrow$ \\
Avatar--Content Pointing & Occurrence of directed gestures and accuracy of their targets & Pointing Accuracy / Pointing Rate~$\uparrow$ \\
Visual Quality & Perceptual quality of the avatar and complete visual scene & Visual Quality Score~$\uparrow$ \\
Motion Smoothness & Temporal smoothness of presenter motion & Normalized Jerk~$\downarrow$ \\
\bottomrule
\end{tabularx}
\end{table}

\begin{figure*}[!t]
\centering
\includegraphics[trim=0 103bp 0 115bp,clip,width=\textwidth]{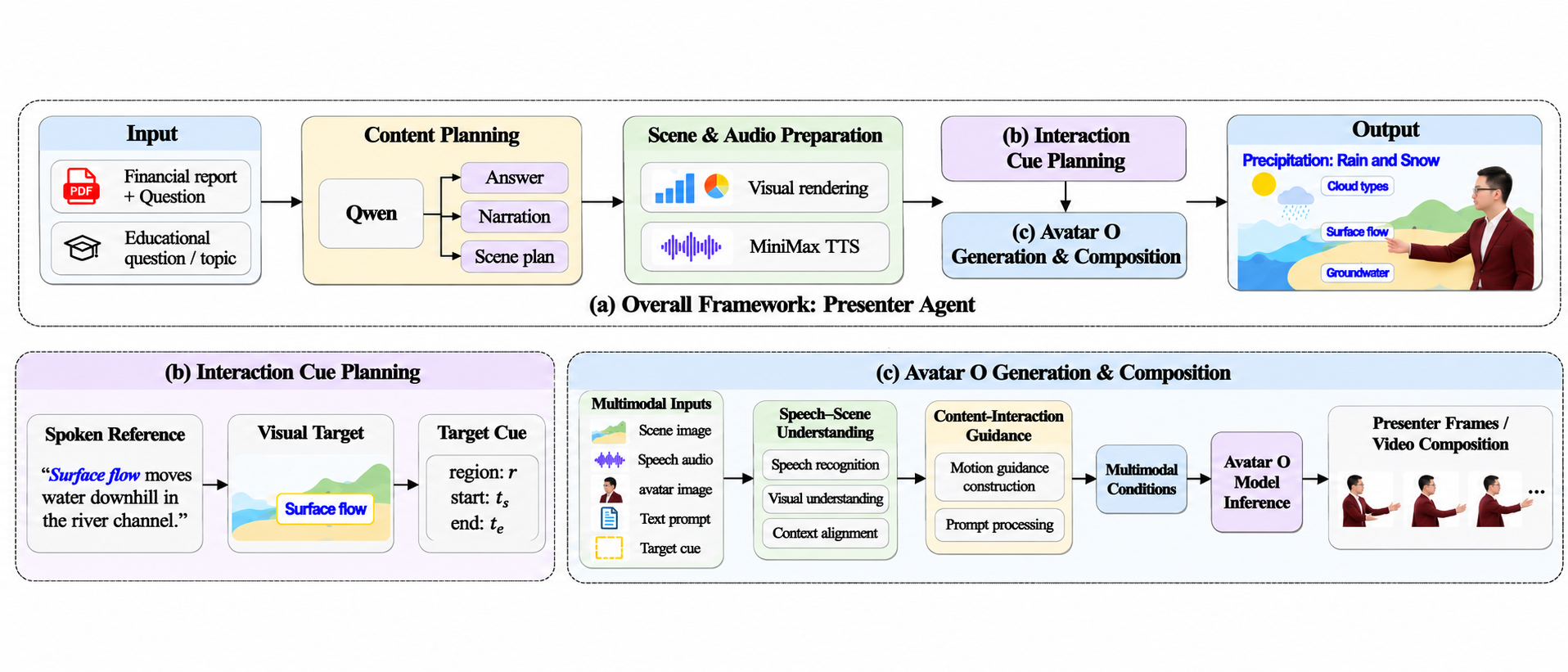}
\caption{Avatar-O Agent. (a) The overall pipeline plans the explanation, prepares visual and speech materials, and generates the presenter video. (b) Interaction planning binds a spoken reference to a rendered target region and time interval. (c) Avatar-O uses the scene image, narration audio, avatar reference image, text prompt, and interaction cues to synthesize presenter frames for final composition.}
\label{fig:avatar-o-inference}
\end{figure*}

\paragraph{Content factuality.}
We extract factual claims made in the delivered narration and visible content~\cite{factscore,tifa,dsg}. An LLM-based evaluator checks them against the relevant reference evidence and independently verifiable knowledge; each claim is categorized as supported, a valid extension, incorrect, or uncertain. A valid extension conveys relevant and independently verifiable information not explicitly stated in the reference. Speech and visual claims are evaluated separately. For modality \(m\), factual precision is
\[
F_m=100\frac{n_{\mathrm{sup},m}+n_{\mathrm{ext},m}}{n_{\mathrm{sup},m}+n_{\mathrm{ext},m}+n_{\mathrm{err},m}}.
\]

Uncertain claims are excluded from the denominator and reported in assessment coverage. This measurement concerns the validity of \textbf{expressed claims}; it does not measure information that the system omitted. Consequently, a high factual-precision score should not be interpreted as proof that an explanation fully answers the request.

\paragraph{Speech--visual consistency.}
For each applicable spoken unit, we inspect the visual material displayed at the corresponding time and assess whether it communicates the same information. Full agreement, partial agreement without a material contradiction, and disagreement or missing relevant visual evidence receive ordinal judgments that are rescaled to a 0--100 score. This dimension is distinct from factual precision: speech and visuals can each be correct yet fail to correspond temporally or semantically. For example, a chart may correctly show two reporting periods while the narration discusses only one. Checking the values for factual support does not resolve whether the quantity visible at the moment of explanation corresponds to the period and series under discussion.

\paragraph{Avatar--Content Pointing.}
We first identify the visual referent of a time-aligned spoken phrase and locate the corresponding region in the system's \textbf{final video frames}, before inspecting the avatar's hand or arm. The evaluator does not use the generator's planned target boxes or gesture instructions. Only a uniquely identifiable visual target permits a spatial pointing judgment. A directed gesture is labeled \emph{Hit} when it identifies the intended region, \emph{Near} when it unambiguously indicates that region despite a small spatial deviation into adjacent empty space, and \emph{Miss} when it indicates another region or fails to identify the target. We separately record \emph{Not pointing} when an assessable reference has no directed gesture, and \emph{Unknown} when the speech target, motion, or visual evidence cannot be resolved. Near is a categorical assessment of unambiguous reference, not a claimed universal pixel-distance threshold. This distinction is particularly important for dense charts. A gesture just outside a bar can still identify it if the neighboring space is empty. A gesture at a neighboring bar can instead select a different series, even when the physical displacement is small.

Let \(H,N,M,P\) denote counts of hits, near hits, misses, and assessable references without a directed gesture, respectively. We report
\[
A_p=100\frac{H+0.5N}{H+N+M},\qquad
R_p=100\frac{H+N+M}{H+N+M+P}.
\]

Pointing Accuracy \(A_p\) evaluates the target of gestures that occurred, while Pointing Rate \(R_p\) measures their occurrence during assessable spoken references. Unknown and inapplicable cases are excluded from both denominators and reported separately. When \(H+N+M=0\) and \(P>0\), Pointing Rate is zero and Pointing Accuracy is undefined. When \(H+N+M+P=0\), both metrics are undefined. The supporting record retains the relevant utterance, time interval, visual target, and observed-frame evidence. Evaluation uses sampled frames at 5 fps, with additional inspection of candidate hits and near hits.

\paragraph{Visual quality.}
We assess the perceptual quality of the complete generated video, considering both the avatar and the surrounding visual scene. The evaluation covers visible artifacts and distortions affecting the presenter's face, hands, and body, as well as the background and other rendered visual elements~\cite{thqa,evaltalker}. We report a severity-weighted quality score over assessable video intervals and their duration coverage. Visual factual precision separately measures the correctness of facts conveyed on screen.

\paragraph{Motion smoothness.}
Motion smoothness is characterized using the image-width-normalized third temporal derivative of tracked pose positions (jerk)~\cite{flashhogan}. Jerk is a motion descriptor, not a standalone measure of naturalness: nearly stationary presenters can have low jerk while providing little useful guidance. We therefore interpret it alongside observed movement and pose-detection coverage.

\paragraph{Aggregation and coverage.}
Scene-level observations are first combined within each scene and then across the scenes belonging to a request. In the financial collection, requests are averaged within their reports before the 16 reports receive equal weight; educational requests receive equal weight. Section~\ref{sec:experiments} presents the available system-level aggregates. Generation failures, evaluator failures, unknown judgments, and inapplicable events are distinguished; an empty denominator produces no score. This procedure does not produce a single overall benchmark ranking.

\section{Avatar-O Agent}
\label{sec:systems}

Section~\ref{sec:task} introduced Avatar-O Agent as the reference implementation used to construct the curated reference-video corpus. We evaluate the same agent alongside five others on the independent evaluation set (Section~\ref{sec:experiments}). The VisualTalk-Bench task and evaluation protocol are architecture-agnostic.

Figure~\ref{fig:avatar-o-inference} shows the complete pipeline. Given a question, an optional source document, a full-body avatar reference image, and a target language, Avatar-O Agent first uses Qwen to generate an answer, a narration script, and a scene plan. Based on the scene plan, the agent renders the corresponding scene images and synthesizes narration audio using MiniMax TTS. These outputs provide the visual and speech content for subsequent avatar generation.

To establish interactions between the avatar and the visual content, the agent identifies spoken references in the narration script and aligns them with visual targets in the scene images. Each target is associated with a spatial region and a temporal interval, forming a pointing cue. The generation module then performs speech--scene understanding and constructs content-interaction guidance from these cues. The resulting guidance, together with the scene image, narration audio, avatar reference image, and text prompt, conditions the post-trained Avatar-O model to generate presenter frames (Figure~\ref{fig:avatar-o-inference}c). Finally, the agent composites the presenter frames with the scene image and narration audio to produce each scene video. For requests involving multiple scenes, the scene videos are assembled in sequence to form the final video explanation.

\subsection{Content and Asset Preparation}
Qwen plans a narrated answer from the educational question or from passages retrieved by lexical matching in the supplied report pages. Financial plans are checked against quantities, units, and reporting periods; educational plans are checked for consistency between concepts and examples. The visual planner selects a chart or diagram for each scene. As the browser renders it, the agent records labels, values, and pixel regions in the layout the viewer will see.

MiniMax TTS synthesizes the narration in speech units. Their durations define phrase boundaries in the assembled audio. When a backend requires a scene to be split, the agent preserves narration order and computes cue times relative to the resulting parts.

\subsection{Interaction Planning and Video Synthesis}
\noindent\textbf{Presenter-oriented video adaptation.} Avatar-O, the video-generation model within Avatar-O Agent, builds on a pretrained video generator. Post-training on human-action data adapts the model to the timing and expressive characteristics of presenter motion.

At inference time, the agent resolves each reference using the labels, text, values, and units of rendered objects. It issues a pointing instruction only when the reference identifies a unique object. For video part \(p\), the instructions are
\[
\mathcal C_p=\bigl\{(u_i,b_i,[t_i^{\mathrm s},t_i^{\mathrm e}])\bigr\}_{i=1}^{N_p}.
\]
Each of the \(N_p\) instructions pairs a spoken phrase \(u_i\) with a target region \(b_i\) in image coordinates. Fixed timing rules use the synthesized speech boundaries to set the gesture's start and end times relative to the start of the part. The spatial and temporal cues come from different stages of preparation: the rendered layout establishes where the object appears, and the synthesized utterance establishes when it is named. Keeping the phrase, region, and interval together allows successive references in the same scene to guide different actions.

To assess the executed gesture, the evaluator independently locates the spoken referent in the final video without access to the generation-time target regions or timing cues.

\section{Experiments}
\label{sec:experiments}

\begin{table*}[t]
\centering\small
\setlength{\tabcolsep}{3pt}
\caption{Agent-level quantitative results on the fixed evaluation set. Factual precision, speech--visual consistency, visual quality, Pointing Accuracy, and Pointing Rate are on a 0--100 scale. Jerk is measured in image widths/s$^3$; lower values indicate smoother motion but not necessarily better guidance. Pointing Accuracy is conditioned on assessable directed gestures; Pointing Rate records their frequency among assessable references. N/A indicates that Pointing Accuracy is undefined when no assessable directed pointing gestures occur. Pointing Rate is zero when assessable spoken references exist but none elicit a directed pointing gesture. General non-pointing gestures are not counted as successful pointing events.}
\label{tab:results}
\begin{tabular*}{\textwidth}{@{\extracolsep{\fill}}lrrrrrrr@{}}
\toprule
Agent configuration & \shortstack[r]{Speech factual\\precision $\uparrow$} & \shortstack[r]{Visual factual\\precision $\uparrow$} & \shortstack[r]{Speech--visual\\consistency $\uparrow$} & \shortstack[r]{Visual\\Quality $\uparrow$} & Jerk $\downarrow$ & \shortstack[r]{Pointing\\Accuracy $\uparrow$} & \shortstack[r]{Pointing\\Rate}\\
\midrule
\textbf{Avatar-O Agent (ours)} & 98.68 & \textbf{99.51} & 95.87 & 80.76 & 50.27 & \textbf{49.85} & \textbf{95.39}\\
MiniMax-backed Agent & 96.61 & 97.13 & 88.08 & 61.26 & 52.71 & 4.63 & 68.32\\
Seedance-backed Agent & 96.12 & 94.36 & 74.17 & 67.55 & 41.86 & 21.15 & 82.74\\
HeyGen-backed Agent & 98.78 & 99.12 & \textbf{96.13} & \textbf{84.75} & 52.52 & N/A & 0.00\\
LiveAct-backed Agent & 98.63 & 96.68 & 90.76 & 24.88 & 57.43 & N/A & 0.00\\
Vidu-backed Agent & \textbf{99.30} & 98.20 & 93.10 & 70.09 & \textbf{30.87} & N/A & 0.00\\
\bottomrule
\end{tabular*}
\end{table*}

\subsection{Experimental Setup}

\noindent\textbf{Comparison Protocol.} We evaluate six complete agent configurations on the same evaluation requests. Each agent receives the same question, optional source document, full-body avatar reference image, and target language. Each agent implements its own pipeline for content preparation and video generation, while following the same scene requirements. The generated scene videos are assembled into a final video explanation, which is evaluated using the same output-based protocol. Thus, our comparison measures the performance of complete task-to-video agents rather than isolated video-generation models.

\noindent\textbf{Baseline Selection.} We select two complementary categories of agents: general-purpose video-generation agents (MiniMax-backed Agent and Seedance-backed Agent) and avatar-oriented agents (HeyGen-backed Agent, LiveAct-backed Agent, and Vidu-backed Agent), alongside our Avatar-O Agent. General-purpose agents generate explanatory videos from LLM-generated explanation content, whereas avatar-oriented agents rely on LLM-generated narration scripts and scene images to prepare the visual content before presenter-video generation. Comparing these two approaches allows us to examine differences in content consistency, visual quality, and avatar--content interaction. A recent study found that embodied pedagogical agents achieved stronger learning outcomes than simple visual cues, even though color cues attracted greater attention to relevant content~\cite{li2026embodied}. This finding further motivates the evaluation of presenter--content interaction.

\noindent\textbf{Avatar-O Agent (Ours).} Our modular reference agent integrates content planning, scene and audio preparation, and explicit avatar--content interaction guidance. It uses the post-trained Avatar-O model for presenter generation. The complete generation pipeline is described in Section~\ref{sec:systems}.

\noindent\textbf{MiniMax-backed Agent.} An LLM first generates the explanation content from the question and optional source document. The agent then uses its built-in chart reference template, together with the explanation content and full-body avatar reference image, to guide MiniMax-H3 in generating each scene video. MiniMax-H3 operates in \texttt{ref2va} mode through SGLang, using 24 inference steps and \texttt{flow\_shift=12} for all evaluation scenes.

\noindent\textbf{Seedance-backed Agent.} An LLM first generates the explanation content from the question and optional source document. The agent then uses its built-in chart reference template, together with the explanation content and full-body avatar reference image, to guide Seedance 2.5 Flash in generating each scene video.

\noindent\textbf{HeyGen-backed Agent.} An LLM generates the explanation content and prepares the corresponding narration script and scene image. Since HeyGen Avatar IV does not independently generate the required explanatory visualizations in this configuration, the agent prepares the scene image before video generation. HeyGen Avatar IV receives the scene image and narration script through its image-driven interface, synthesizes the narration audio, and generates each scene video.

\noindent\textbf{LiveAct-backed Agent.} An LLM generates the explanation content and prepares the corresponding narration script and scene image. Since SoulX-LiveAct 18B does not independently generate the required explanatory visualizations in this configuration, the agent prepares the scene image before video generation. The agent then uses SoulX-LiveAct 18B to generate the presenter video and combines it with the scene image to produce each scene video.

\noindent\textbf{Vidu-backed Agent.} An LLM generates the explanation content and prepares the corresponding narration script and scene image. Since Vidu S2 Avatar does not independently generate the required explanatory visualizations in this configuration, the agent prepares the scene image before video generation. The agent then uses the scene image and narration script to prepare the supported generation inputs for Vidu S2 Avatar, which generates each scene video.

\subsection{Quantitative Results}

Table~\ref{tab:results} reports the observed results for six agent configurations on the same fixed set of 100 requests and 140 scenes per configuration. Full video coverage does not imply an equal number of \textbf{eligible events} for every metric: pointing in particular depends on whether a spoken target and observed gesture can be assessed. For example, an assessable passage in which the presenter never points contributes to the rate denominator. A passage in which the intended object cannot be resolved is excluded from both pointing denominators. 

\noindent\textbf{Content factuality and speech--visual consistency.} Speech factual precision ranges from 96.12 to 99.30 and visual factual precision from 94.36 to 99.51. These scores concern expressed claims and reflect the content preparation and video generation of each complete agent. The Seedance-backed Agent achieves 96.12 and 94.36 on the two measures, yet its speech--visual consistency is 74.17. A narration and a chart can each contain valid information while presenting different facts at the same moment.

\noindent\textbf{Visual quality and guidance.} The HeyGen-backed Agent achieves the highest visual quality (84.75) and speech--visual consistency (96.13), compared with 80.76 and 95.87 for Avatar-O Agent. The Vidu-backed Agent has the lowest jerk (30.87), which can also occur when a presenter barely moves. These outcomes do not establish whether a hand indicates the correct object. For the HeyGen-backed Agent, LiveAct-backed Agent, and Vidu-backed Agent, assessable references occur without assessable directed pointing: Pointing Rate is 0.00 and Pointing Accuracy is N/A. These agents do not implement content-aware pointing; their backends may generate general gestures.

\noindent\textbf{Pointing rate and accuracy.} Avatar-O Agent achieves a Pointing Rate of 95.39 and a Pointing Accuracy of 49.85. The Seedance-backed Agent also points frequently (82.74) but has a Pointing Accuracy of 21.15; the MiniMax-backed Agent reaches 68.32 and 4.63, respectively. Thus a directed gesture can accompany a phrase without identifying its referent. For Avatar-O Agent, the gap is especially informative: a gesture accompanies nearly every assessable reference, yet the weighted accuracy of those gestures is below 50. Increasing gesture frequency alone would not correct an error in which a hand repeatedly indicates an adjacent chart element. These descriptive differences apply to complete configurations, and small margins do not establish reliable advantages without paired uncertainty estimates.

\noindent\textbf{Implications for Agent Design.} Our findings highlight the need to coordinate spoken references, visual targets, and presenter actions in avatar-led explanations. One promising direction is to resolve the visual referent of each spoken phrase, associate it with the rendered scene, and align gesture generation with the corresponding spatial and temporal cues. Avatar-O Agent explores this approach through explicit interaction planning, yet its remaining pointing errors show that planned targets do not guarantee accurate execution. Future systems may therefore benefit from stronger referent resolution, spatially grounded motion generation, and verification that observed pointing gestures identify the intended visual targets. These capabilities could be implemented through modular pipelines or learned jointly by end-to-end models.

\subsection{Human Evaluation}
\label{sec:human-evaluation}

\noindent\textbf{Study protocol.} To complement the automatic evaluation, we conducted a human study comparing four representative agent configurations: Avatar-O Agent, MiniMax-backed Agent, Seedance-backed Agent, and HeyGen-backed Agent. The 48 requests were divided among four questionnaires. Each contained four financial, four Chinese educational, and four English educational requests. Participants evaluated one video from each agent under anonymized labels for each request, yielding 48 video evaluations per completed questionnaire. Each video was rated on three five-point Likert scales: \textbf{visual guidance}, \textbf{perceived pointing accuracy}, and \textbf{explanation effectiveness}. Participants also selected their overall preferred explanation for each request, with options for no clear difference and inability to judge.

We collected responses from 26 unique participants, 22 through Feishu and four offline. Eight participants completed all 12 requests in their assigned questionnaire, while 18 provided partial responses. We retained 2,020 valid item-level ratings and 185 request-level preference responses. The 2,020 ratings are repeated measurements from the same 26 participants across videos and rating dimensions. We report descriptive mean scores and preference frequencies in Table~\ref{tab:human_eval}.

\begin{table}[t]
\centering
\caption{Human evaluation of four agent configurations. All ratings use a five-point scale (higher is better). Guide.: visual guidance; Point.: perceived pointing accuracy; Explain.: explanation effectiveness. Preference percentages use 181 judgeable responses, including 10 ties; four cannot-judge responses are excluded.}
\label{tab:human_eval}
\small
\setlength{\tabcolsep}{3pt}
\begin{tabular}{@{}lcccc@{}}
\toprule
Agent & Guide. & Point. & Explain. & Pref. \\
\midrule
Avatar-O Agent (ours) & \textbf{3.59} & \textbf{3.62} & \textbf{3.70} & \textbf{76 (42.0\%)} \\
Seedance-backed Agent & 2.94 & 3.03 & 3.21 & 30 (16.6\%) \\
MiniMax-backed Agent & 2.79 & 2.93 & 3.09 & 31 (17.1\%) \\
HeyGen-backed Agent & 2.05 & 2.04 & 3.42 & 34 (18.8\%) \\
\bottomrule
\end{tabular}
\end{table}

\noindent\textbf{Human evaluation results.} Avatar-O Agent achieves the highest mean score on all three subjective dimensions, with 3.59 for visual guidance, 3.62 for perceived pointing accuracy, and 3.70 for explanation effectiveness. It also receives 76 of the 181 judgeable preference responses (42.0\%), compared with 34 for the HeyGen-backed Agent, 30 for the Seedance-backed Agent, and 31 for the MiniMax-backed Agent. Ten responses indicate no clear difference, while four additional responses are excluded from preference percentages because participants could not make a judgment.

The HeyGen-backed Agent ranks second on explanation effectiveness (3.42), despite lower ratings for visual guidance (2.05) and perceived pointing accuracy (2.04). Viewers can find a video explanatory even when they perceive little object-level guidance. These ratings and the automatic metrics have different observation units. A viewer can judge how clearly a whole video directs attention even when it contains no assessable directed gesture. Thus the HeyGen-backed Agent can receive a subjective pointing rating although its automatic accuracy is undefined. The viewer scores describe perceived guidance; the event-level measure asks whether an observed gesture identifies a particular spoken referent. The preference question concerns the explanation as a whole. A viewer who favors one video may respond to its narration, visual material, or presenter behavior, and the choice alone cannot identify which of these influenced the judgment.

\noindent\textbf{Interpretation and limitations.} The participant pool is small and includes partial questionnaires. The comparisons are descriptive; no significance tests were conducted. Viewers judged complete videos, so the study neither isolates pointing against highlighting alone nor establishes improvements in audience learning.

\section{Conclusion}
We introduced VisualTalk-Bench to evaluate question-driven avatar-led video explanations, particularly the alignment between presenter gestures and visual content. Our evaluation reveals different challenges across generation approaches. General-purpose video-generation agents struggle with precise pointing, while the evaluated avatar-oriented agents rely on separately prepared visual content and lack explicit content-aware pointing. Avatar-O Agent demonstrates the potential of interaction planning, although its Pointing Accuracy remains limited. These findings suggest two complementary research directions. \textbf{Agent-based approaches} could improve visual target grounding, speech--gesture alignment, and verification of generated actions. \textbf{End-to-end multimodal models} could jointly understand questions, narration, visual content, and avatar actions, enabling more coordinated explanation and video generation. Beyond these technical directions, VisualTalk-Bench could be extended to broader domains and languages, dynamically updated visualizations, and real-time interactions in which avatars respond to user questions and changes in visual content.

\section*{LLM Usage Statement}
OpenAI ChatGPT and Codex were used to assist with research planning, related-work analysis, manuscript drafting and revision, and LaTeX editing. OpenAI image-generation tools assisted with figure preparation and editing. In our research pipeline, Qwen (qwen3.6-flash) was used for content preparation and explanation planning, while MiniMax TTS (speech-2.8-hd) was used for narration synthesis. LLM-based methods were also employed for content quality screening and factuality evaluation. The video-generation models used in our experiments are detailed in Sections~\ref{sec:systems} and~\ref{sec:experiments}. The authors remain responsible for the accuracy of the research, experimental results, references, and final manuscript.
\pdfbookmark[1]{References}{visualbench-references}
{\small
\bibliographystyle{ieeenat_fullname}
\nocite{syncnet}
\bibliography{references}
}
\end{document}